\documentclass[11pt,a4paper]{article}
\usepackage[hyperref]{emnlp2018}
\usepackage{times}
\usepackage{amsmath}
\usepackage{amsfonts}
\usepackage{latexsym}
\usepackage{amssymb}
\usepackage{url}
\usepackage{bm}

\usepackage[colorinlistoftodos]{todonotes}

\usepackage{algpseudocode}
\usepackage{algorithm}
\usepackage{multirow}
\usepackage{subfigure}
\usepackage{arydshln}
\usepackage{booktabs}

\aclfinalcopy 

\newcommand{\specialcell}[2][c]{%
  \begin{tabular}[#1]{@{}l@{}}#2\end{tabular}}

\title{Multi-Source Domain Adaptation with Mixture of Experts}

\author{Jiang Guo, Darsh J Shah and Regina Barzilay \\
  Computer Science and Artificial Intelligence Laboratory\\
  Massachusetts Institute of Technology \\
  {\tt \{jiang\_guo, darsh, regina\}@csail.mit.edu} \\}
  
\date{}

\begin{document}
\maketitle
\begin{abstract}

We propose a mixture-of-experts approach for unsupervised domain adaptation from multiple sources. The key idea is to explicitly capture the relationship between a target example and different source domains. This relationship, expressed by a \textit{point-to-set} metric, determines how to combine predictors trained on various domains. The metric is learned in an unsupervised fashion using \textit{meta-training}. Experimental results on sentiment analysis and part-of-speech tagging demonstrate that our approach consistently outperforms multiple baselines and can robustly handle negative transfer.\footnote{Our code and data are available at \url{https://github.com/jiangfeng1124/transfer}.} 
\end{abstract}

\section{Introduction}
\label{sec:intro}

Typical domain adaptation methods are designed to transfer supervision from a single source domain.
However, in many practical applications, we have access to multiple sources. For instance, in sentiment analysis of product reviews, we can often transfer from a wide range of product domains, rather than one. This can be particularly promising for target domains which do not match any one available source well. For example, the \textit{Kitchen} product domain may include reviews on \textit{pans}, \textit{cookbooks} or \textit{electronic devices}, which cannot be perfectly aligned to a single source such as \textit{Cookware}, \textit{Books} or \textit{Electronics}. 
By intelligently aggregating distinct and complementary information from multiple sources, we may be able to better fit the target distribution.

A straightforward approach to utilizing data from multiple sources is to combine them into a single domain.
This strategy, however, does not account for distinct relations between individual sources and the target example. Constructing a common feature space for this heterogeneous collection may wash out informative characteristics of individual domains and also lead to \textit{negative transfer} \cite{rosenstein2005transfer}.

Therefore, we propose to explicitly model the relationship between different source domains and target examples. We hypothesize that different source domains are aligned to different sub-spaces of the target domain.
Specifically, in this paper, we model the domain relationship with a mixture-of-experts (MoE) approach \cite{jacobs1991adaptive}. For each target example, the predicted posterior is a weighted combination of all the experts' predictions. The weights reflect the proximity of the example to each source domain. Our model learns this \textit{point-to-set} metric automatically, without additional supervision.


We define the \textit{point-to-set} metric using Mahalanobis distance \cite{weinberger2009distance} between individual examples and a set (i.e. domain), which are computed within the hidden representation space of our model. The main challenge is to learn this metric in an unsupervised setting. We address it through a \textit{meta-training} procedure, in which we create multiple \textit{meta}-tasks of domain adaptation from the source domains.
In each \textit{meta}-task, we pick one of the source domains as \textit{meta-target}, and the rest source domains as \textit{meta-sources}. By minimizing the loss using the MoE predictions on \textit{meta-target}, we are able to learn both the model and the metric simultaneously. To further improve transfer quality, we align the encoding space of our target and source domains via adversarial learning.

We evaluate our approach on sentiment analysis using the benchmark multi-domain Amazon reviews dataset \cite{Chen2012msda,ziser2017nscl} as well as on part-of-speech (POS) tagging using the SANCL dataset \cite{petrov2012sancl}.
Experiments show that our approach consistently improves the adaptation results over the best single-source model and a unified multi-source model. On average, we achieve a 7\% relative error reduction on the Amazon reviews dataset, and a 13\% on the SANCL dataset. Importantly, the POS tagging experiments on the SANCL dataset demonstrate that our method is able to robustly handle negative transfer from unrelated sources (e.g., \textit{Twitter}) and utilize it effectively to consistently improve performance.



\section{Related Work}
\label{sec:relwork}
\paragraph{Unsupervised domain adaptation} 
Most existing domain adaptation methods focus on aligning the feature space between source and target domains to reduce the domain shift \cite{ben2007analysis,blitzer2007da,blitzer2006domain,pan2010cross}.
Our approach is close to the representation learning approaches, such as the denoising autoencoder \cite{glorot2011domain}, the marginalized stacked denoising autoencoders \cite{Chen2012msda}, and domain adversarial networks \cite{tzeng2014deep,ganin2016domain,zhang2017aspect,Shen2018WassersteinDG}. 

In contrast to these previous approaches, however, our approach not only learns a shared representation space that generalizes well to the target domain, but also captures informative characteristics of individual source domains.



\paragraph{Multi-Source domain adaptation}
The main challenge in using multiple sources for domain adaptation is in learning domain relations. Some approaches assume that all source domains are equally important to the target domain \cite{li2008multi, luo2008transfer, crammer2008learning}. Others learn a global domain similarity metric using labeled data in a supervised fashion \cite{yang2007cross,duan2009domain,yu2018meta} or use predefined similarity measures \cite{ruder2017knowledge}. Alternatively, \newcite{mansour2009domain} and \newcite{bhatt2016cross} utilize unlabeled data of the target domain to find a distribution weighted combination of the source domains or to construct an auxiliary training set of the source domain instances close to the target domain instances. Recent adversarial methods on multi-source domain adaptation \cite{zhao2018multiple,chen2018multinomial} align source domains to the target domains globally, without accounting for the distinct importance of each source with respect to a specific target example.

Beyond the global \textit{domain-to-domain} relations, \newcite{ruder2017transfer} use \textit{example-to-domain} similarity measures to select data from multiple sources for a specific target domain, which are then combined to train a single classifier. The work most related to ours is by \newcite{kim2017domain}. They also model the \textit{example-to-domain} relations, but use an attention mechanism. The attention module is learned using limited training data from the target domain in a supervised fashion. Our method, however, works in an unsupervised setting without utilizing any labeled data from the target domain.



%


\section{Methodology}
\label{sec:method}

\paragraph{Problem definition}
We follow the unsupervised multi-source domain adaptation setup, assuming access to labeled training data from $K$ source domains: $\{\mathcal{S}_i\}_{i=1}^{K}$ where $\mathcal{S}_i\triangleq\{(x_t^{\mathcal{S}_i}, y_t^{\mathcal{S}_i})\}_{t=1}^{\left|\mathcal{S}_i\right|}$, and (optionally) unlabeled data from a target domain: $\mathcal{T}\triangleq\{x_t^{\mathcal{T}}\}_{t=1}^{\left|\mathcal{T}\right|}$.
The goal is to learn a model using the source domain data, that generalizes well to the target domain.

\paragraph{Notations} For the rest of the paper, we denote an individual example as $x$, and a batch of examples as $\mathbf{x}$. We use superscript to denote the domain from which an example is sampled, and use subscript to denote the index of an example.

\subsection{Overview of Our Approach}

\begin{figure}
    \centering
    \includegraphics[width=76mm]{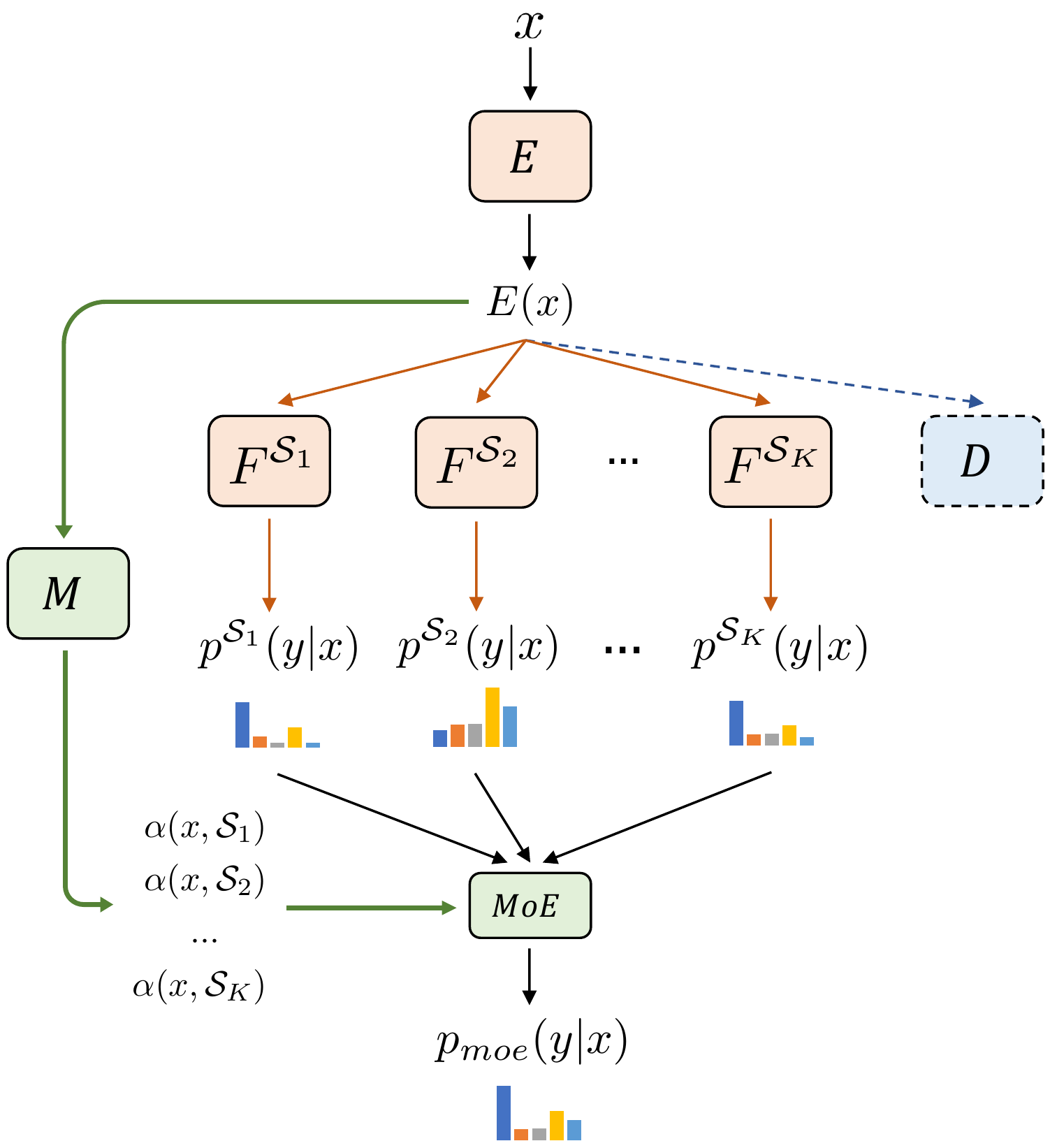}
    \caption{Architecture of the MoE model. $E$ is the encoder which maps an input $x$ to a hidden representation $E(x)$; $F^{\mathcal{S}_i}$ is the classifier on the $i^{th}$ source domain; $D$ is the critic that is only used during adversarial training. $M$ is the metric learning component, which takes the encoding of $x$ and source domains ($\mathcal{S}_{1:K}$) as input and computes $\alpha$.}
    \label{fig:moe}
\end{figure}

We model the multiple source domains as a mixture of experts, and learn a \textit{point-to-set} metric  $\alpha$ to weight the experts for different target examples. The metric is learned in an unsupervised manner.

Our model consists of four key components as shown in Figure \ref{fig:moe}, namely the \textit{encoder} ($E$), \textit{classifier} ($F$), \textit{metric} ($M$) and \textit{adversary} ($D$). We use a typical neural multi-task learning architecture \cite{caruana1997multitask}, with a shared encoder across all sources, and domain-specific classifiers ($\{F^{\mathcal{S}_i}\}_{i=1}^K$). Each input is first encoded with $E$, and then fed to each classifier to obtain the domain-specific predictions (i.e. posteriors). The final predictions are then weighted based on the \textit{metric} (see Equation \ref{eq:1}).

We start by describing the representation learning component.



\subsection{Representation}
\label{sec:repr}




Our goal is to design an encoder that supports transfer, while maintaining source domain-specific information. Depending on different tasks and datasets, we select appropriate encoders --- \textit{multilayer perceptron} (MLP), \textit{convolutional neural network} (CNN) or \textit{long short-term memory networks} (LSTM) (see Section \ref{subsec:implement} for details).

We further add an adversarial module ($D$) on top of the encoder, in order to align the target domain with the sources. $D$ is typically designed as a parameterized classifier in domain adversarial networks \cite{ganin2016domain,zhang2017aspect}, which is trained jointly with the encoder and the classifiers through a minimax game. Here, we instead use Maximum Mean Discrepancy (MMD) \cite{gretton2012kernel} as our adversary. This distance metric measures the discrepancy between two distributions explicitly in a non-parametric manner, greatly simplifying the training procedure compared to domain adversarial networks which use an additional domain classifier module.

\subsection{Mixture of Experts}
\label{sec:moe}
Given an example $x$ from the target domain, we model its posterior distribution as a mixture of posteriors produced by models trained on different source domain data:
\begin{equation}
\begin{split}
    p_{moe}(y|x) & = \sum_{i=1}^{K} \alpha (x, \mathcal{S}_i)\cdot p^{\mathcal{S}_i}(y|x) \\
                 & = \sum_{i=1}^{K} \alpha (x, \mathcal{S}_i)\cdot \mathtt{softmax}\big(\mathbf{W}^{\mathcal{S}_i} E(x)\big)
\end{split}
\label{eq:1}
\end{equation}
$p^{\mathcal{S}_i}$ is the posterior distribution produced by the $i^{th}$ source classifier $F^{\mathcal{S}_i}$ (the $i^{th}$ expert). $\mathbf{W}^{\mathcal{S}_i}$ is the output layer weights of $F^{\mathcal{S}_i}$, $\alpha$ is a parameterized metric function that measures how much confidence we put in the specific source expert for a given example $x$.\footnote{In typical MoE frameworks \cite{jacobs1991task,jacobs1991adaptive,shazeer2017outrageously}, $\alpha$ is commonly realized as a ``gating network", which produces a normalized weight vector that determines the combination of experts depending solely on the input example. Such gating networks, however, do not yield promising results in our scenario. We hypothesize that both the input example and the underlying domain distribution should be captured for determining the credit assignment.}
To derive $\alpha$, we first define a \textit{point-to-set} Mahalanobis distance metric between an example $x$ and a set $\mathcal{S}$:
\begin{equation*}
    d(x, \mathcal{S}) = \Big( \big(E(x) - \mu^\mathcal{S}\big)^\top \mathbf{M}^{\mathcal{S}}\big(E(x) - \mu^\mathcal{S}\big) \Big)^{\frac{1}{2}}
\end{equation*}
where $\mu^\mathcal{S}$ is the mean encoding of $\mathcal{S}$. In its original form, the matrix $\mathbf{M}^\mathcal{S}$ played the role of the inverse covariance matrix. However, computing the inverse of the covariance matrix is both time consuming and numerically unstable in practice. Here we allow $\mathbf{M}$ to denote any positive semi-definite matrix which is to be estimated during training \cite{weinberger2009distance}.
To guarantee the positive semi-definiteness of $\mathbf{M}$, we approximate $\mathbf{M}$ with $\mathbf{M} = \mathbf{U} \mathbf{U}^\top$, where $\mathbf{U}\in \mathbb{R}^{h\times r}$, $h$ is the dimension of hidden representations and $r$ is a hyper-parameter controlling the rank of $\mathbf{M}$.

Based on the distance metric, we further derive a confidence score $e(x, \mathcal{S}_i)=f\big(d(x, \mathcal{S}_i)\big)$ for each specific expert.
The final metric values $\alpha(x, \mathcal{S}_i)$ are then obtained by normalizing these confidence scores:
\begin{equation}
    \alpha (x, \mathcal{S}_i) = \frac{\exp\big(e(x, \mathcal{S}_i)\big)}{\sum_{j=1}^{K}\exp\big(e(x, \mathcal{S}_j)\big)}
\end{equation}

Here, we explain our design of $e(x, \mathcal{S})$ on two tasks, respectively \textit{binary classification} and \textit{sequence tagging}, which are also used for evaluation in this paper (Section \ref{sec:exp-setup}).
\paragraph{Binary classification} The \textit{point-to-set} Mahalanobis distance metric measures the distance between an example $x$ and the mean encoding of $\mathcal{S}$, i.e. $\mu^\mathcal{S}$, while taking into account the (pseudo) covariance of $\mathcal{S}$. In binary classification, however, the mean vector $\mu^\mathcal{S}$ is likely to be located near the decision boundary, particularly under a balanced setting. Therefore, a small $d(x, \mathcal{S})$ actually implies lower confidence of the corresponding classifier, which is counter-intuitive.
To this end, we instead define the confidence $e(x, \mathcal{S})$ as the difference between the distances from $x$ to each category of $\mathcal{S}$, referred to as \textit{Maximum Cluster Difference} in \newcite{ruder2017knowledge}:
\begin{equation*}
    e(x, \mathcal{S}) = \left|d(x, \mathcal{S}^{+}) - d(x, \mathcal{S}^{-})\right|
\end{equation*}
Here $\mathcal{S}^+$ and $\mathcal{S}^-$ stand for the positive space and negative space of $\mathcal{S}$ respectively.
Consequently, if $x$ is either far away from $\mathcal{S}$ (i.e., $x$ is not in the manifold of $\mathcal{S}$) or near the classification boundary, we will get a small $e(x, \mathcal{S})$ indicating a low confidence to the corresponding prediction. On the contrary, if $x$ is much closer to a specific category of $\mathcal{S}$ than other categories, the classifier will get a higher confidence.

\paragraph{Sequence tagging} For sequence tagging tasks (e.g., POS tagging), we compute the distance metric at the \textit{token level}.\footnote{This actually makes it a multi-class classification problem with respect to every token of a sequence.} Unlike in \textit{binary classification}, the decision boundary here is more complicated, and the label distribution is typically imbalanced. The mean vector $\mu^\mathcal{S}$ is unlikely to be located at the decision boundary. So we directly use the (reverse) distance as the confidence value for each token $x$: 
\begin{equation*}
    e(x, \mathcal{S}) = -d(x, \mathcal{S})
\end{equation*}


\subsection{Training}
\label{subsec:train}
Since we do not have annotated data in the target domain, we have to learn our model in an unsupervised fashion. Inspired by the recent progress on few-shot learning with metric-based models such as matching network \cite{vinyals2016matching,yu2018meta} and prototypical network \cite{snell2017prototypical}, we propose the following \textit{meta-training} approach. Given $K$ source domains, each source domain will be considered as a target, referred to as \textbf{\textit{meta-target}}, with the rest of the source domains as \textbf{\textit{meta-sources}}.
This way, we obtain $K$ (\textit{meta-sources}, \textit{meta-target}) training pairs for domain adaptation.
Then, we apply our MoE formulation over these \textit{meta-training} pairs to learn the metric.
At testing time, the metric will be applied to all the $K$ source domains for each example in the target domain.

We optimize two main objectives: the MoE objective and the multi-task learning (MTL) objective.

\paragraph{MoE objective}
For each example in each \textit{meta-target} domain, we compute its MoE posterior using the corresponding \textit{meta-sources}. Therefore, we get the following MoE loss over the entire multi-source training data:
\begin{equation}
\begin{split}
    \mathcal{L}_{moe} & = -\sum_{i=1}^{K}\sum_{j=1}^{\left|\mathcal{S}_i\right|}\log p_{moe}(y^{\mathcal{S}_i}_{j}|x^{\mathcal{S}_i}_{j}) \\
            & = -\sum_{i=1}^{K}\sum_{j=1}^{\left|\mathcal{S}_i\right|}\log \sum_{l=1, l\ne i}^{K}\alpha(x, \mathcal{S}_l)\cdot p^{\mathcal{S}_l}(y^{\mathcal{S}_i}_{j}|x^{\mathcal{S}_i}_{j})
\end{split}
\end{equation}
Note that $\alpha$ is normalized over the \textit{meta-sources} for each \textit{meta-target}, rather than over all the $K$ sources.

\paragraph{MTL objective}
For each \textit{meta-target}, we further optimize a supervised cross-entropy loss using the corresponding labels. All supervised objectives are optimized jointly with the encoder being shared, resulting in the following multi-task learning objective:
\begin{equation}
    \mathcal{L}_{mtl}=-\sum_{i=1}^{K}\sum_{j=1}^{\left|\mathcal{S}_i\right|}\log p^{\mathcal{S}_i}(y^{\mathcal{S}_i}_{j}|x^{\mathcal{S}_i}_{j})
\end{equation}

\begin{algorithm*}[t]
\label{alg:training}
\caption{Training Procedure}
\begin{algorithmic}[1]
\State \textbf{Input}: multi-source domain data $\mathcal{S} = \{\mathcal{S}_i\}_{i=1}^{K}$, target domain data $\mathcal{T}$
\State \textbf{Hyper-parameters}: mini-batch size $m$, coefficients for different losses: $\lambda$, $\gamma$ and $\eta$
\Repeat
\State Sample $K$ source mini-batches $\{(\mathbf{x}^{\mathcal{S}_i}, \mathbf{y}^{\mathcal{S}_i})\}_{i=1}^{K}$ from $\mathcal{S}$ and a target mini-batch $\mathbf{x}^T$ from $\mathcal{T}$
\State $\mathcal{L}_{mtl}, \mathcal{L}_{moe}, \mathcal{L}_{adv}, \mathcal{R}_{h} \leftarrow 0$
\For{$t = 1$ to $K$}
\State Set \textbf{\textit{meta-target}} as $\mathcal{T}^{meta}\triangleq\Tilde{\mathcal{S}}_t\triangleq(\mathbf{x}^{\mathcal{S}_t}, \mathbf{y}^{\mathcal{S}_t})$
\State Set \textbf{\textit{meta-sources}} as $\mathcal{S}^{meta}\triangleq\{\Tilde{\mathcal{S}}_i\}_{i=1, i\ne t}^{K}$, where $\Tilde{\mathcal{S}}_i\triangleq(\mathbf{x}^{\mathcal{S}_i}, \mathbf{y}^{\mathcal{S}_i})$
\State Compute cross-entropy loss over $\mathcal{T}^{meta}$, and add to $\mathcal{L}_{mtl}$
\State Compute Mahalanobis metric $\alpha(x, \mathcal{S}')$ for each $x\in \mathcal{T}^{meta}$ and $\mathcal{S}'\in \mathcal{S}^{meta}$ \Comment{Eq. (2)}
\State Compute MoE loss over ($\mathcal{S}^{meta}$, $\mathcal{T}^{meta}$) using $\bm{\alpha}$, and add to $\mathcal{L}_{moe}$ \Comment{Eq. (3)}
\State Compute entropy of $\bm{\alpha}(x, \cdot)$ for each $x\in\mathcal{T}^{meta}$, and add to $\mathcal{R}_{h}$ \Comment{Eq. (6)}
\EndFor
\State Compute MMD between $\mathbf{x}^\mathcal{T}$ and $\cup_{i=1}^{K}\mathbf{x}^{\mathcal{S}_i}$, and add to $\mathcal{L}_{adv}$ \Comment{Eq. (5)}
\State Update parameters via backpropagating gradients of the total loss $\mathcal{L}$ \Comment{Eq. (7)}
\Until{converge}
\end{algorithmic}
\end{algorithm*}

\paragraph{Adversary-augmented MoE}
We use MMD \cite{gretton2012kernel} as the adversary to minimize the divergence between the marginal distribution of target domain and source domains.
Specifically, at each training epoch, given the $K$ batches $\{\mathbf{x}^{\mathcal{S}_1},\mathbf{x}^{\mathcal{S}_2},...,\mathbf{x}^{\mathcal{S}_{K}}\}$ from all the source domains, we sample a batch (unlabeled) $\mathbf{x}^\mathcal{T}$ from our target domain, and minimize the MMD:
\begin{equation}
    \mathcal{L}_{adv} = \mathtt{MMD}^2( \mathbf{x}^{\mathcal{S}_1} \cup ... \cup \mathbf{x}^{\mathcal{S}_{K}},\mathbf{x}^\mathcal{T})
\end{equation}
where
\begin{equation*}
\begin{split}
    &\mathtt{MMD}(\mathcal{D}^\mathcal{S}, \mathcal{D}^\mathcal{T}) = \\
    &\left\Vert \frac{1}{|\mathcal{D}^\mathcal{S}|}\sum_{x_s \in \mathcal{D}^\mathcal{S}}\phi\big(E(x_s)\big) - \frac{1}{|\mathcal{D}^\mathcal{T}|}\sum_{x_t \in \mathcal{D}^\mathcal{T}}\phi\big(E(x_t)\big)\right\Vert_{\mathcal{H}} 
\end{split}
\end{equation*}
measures the discrepancy between $\mathcal{D}^\mathcal{S}$ and $\mathcal{D}^\mathcal{T}$ based on Reproducing Kernel Hilbert Space (RKHS). $\phi(\cdot)$ is the feature map induced by a universal kernel. We follow \newcite{bousmalis2016domain} and use a linear combination of multiple RBF kernels: $\kappa(\bm{h}_i, \bm{h}_j) = \Sigma_n \exp(-\frac{1}{2\sigma_n}\left\Vert \bm{h}_i - \bm{h}_j\right\Vert^2)$.

\paragraph{Entropy regularization}
In the \textit{meta-training} process, for each example $x$ in \textit{meta-target}, we know exactly from which source $x$ is sampled. This provides additional insight that the $\alpha$ distribution is skewed, which can be utilized as a soft constraint. Therefore, we propose to regularize the entropy of the $\alpha$ distribution over all the sources, rather than \textit{meta-sources}:\footnote{Alternatively, we can directly exploit this supervision and minimize the KL divergence of the $\alpha$ distribution and its ground truth one-hot distribution. In practice, however, we found it beneficial to allow examples from one domain to be attended to different sources. This observation may be attributed to the fact that each domain indeed consists of multiple latent sub-domains.}
\begin{equation*}
    H\big(\bm{\alpha}(x, \cdot)\big) = -\sum_{l=1}^{K}\mathbf{\alpha}(x, \mathcal{S}_l) \cdot \log \mathbf{\alpha}(x, \mathcal{S}_l)
\end{equation*}
\begin{equation}
    \mathcal{R}_h = \sum_{i=1}^{K}\sum_{j=1}^{\left|\mathcal{S}_i\right|}H\big(\bm{\alpha}(x^{\mathcal{S}_i}_{j}, \cdot)\big)
\end{equation}

\paragraph{Joint learning} Our final objective is the weighted combination of each individual component loss:
\begin{equation}
\begin{split}
    \mathcal{L} = &~ \lambda \cdot \mathcal{L}_{moe} + (1 - \lambda) \cdot \mathcal{L}_{mtl} \\
        & + \gamma \cdot \mathcal{L}_{adv} \\
        & + \eta \cdot \mathcal{R}_{h}
\end{split}
\end{equation}
where $\lambda$ controls the balance of the MoE loss and MTL loss. $\gamma$ is set to 0 in \textit{non-adversarial} setting when unlabeled data from the target domain is not provided.
Additionally, it would be straightforward to add an MoE loss for labeled data in the target domain if they are available, thus extending our framework to a setting where we have few-shot target annotations.
The training process is shown in Algorithm 1.

\section{Experimental Setup}
\label{sec:exp-setup}
\subsection{Task and Dataset}

\paragraph{Sentiment classification} We use the multi-domain Amazon reviews dataset \cite{blitzer2007da}, one of the standard benchmark datasets for domain adaptation. It contains reviews on four domains: \textit{Books} (B), \textit{DVDs} (D), \textit{Electronics} (E), and \textit{Kitchen appliances} (K).

We follow the specific experiment settings proposed by \newcite{Chen2012msda} (\textsc{Chen12}) and \newcite{ziser2017nscl} (\textsc{Ziser17}).
\begin{enumerate}
    \item In \textsc{Chen12}, each domain has 2,000 labeled examples for training (1,000 positive and 1,000 negative), and the target test set has 3,000 to 6,000 examples.\footnote{This dataset has been processed by the author to TF-IDF representations, using the 5,000 most frequent unigram and bigram features, thus word order information is not available.}
    \item In \textsc{Ziser17}, each domain also has 2,000 labeled examples (1,000 positive and 1,000 negative), sampled differently from \textsc{Chen12}.
\end{enumerate}

For each dataset, we conduct experiments by selecting the target domain in a round-robin fashion.
Following the protocol in previous work, we use cross-validation over source domains for hyper-parameters selection for each adaptation task \cite{zhao2018multiple}.
When training with an adversary, we use the 2,000 examples training set of the target domain as the unlabeled data in both the settings. In \textsc{Ziser17}, the same data is also used for test, resulting in a transductive setting.


\paragraph{Part-of-Speech tagging} We further consider a sequence tagging task, where the metric is computed over the token-level encodings and multi-class predictions are made at the token (word) level. We use the SANCL dataset \cite{petrov2012sancl} which contains part-of-speech (POS) tagging annotations in 5 web domains: \textit{Emails}, \textit{Weblogs}, \textit{Answers}, \textit{Newsgroups}, and \textit{Reviews}. Among these, \textit{Newsgroups}, \textit{Reviews}, and \textit{Answers} have both a validation and a test set, and are used as target domains. The test set from \textit{Weblogs} and \textit{Emails} are used as individual source domains. The tagging is performed using the Universal POS tagset \cite{petrov2011universal}. We also use \textit{Twitter} \cite{liu2018parsing} as an additional training source. Since it differs substantially from other sources and the target domain, we can assess our model's ability to handle negative transfer. We consider 750 sentences from each SANCL source domain for training, and up to 2,250 sentences from the \textit{Twitter} dataset to magnify the negative transfer. The validation set in the standard split of each target domain is used for hyper-parameters selection and early-stopping in our experiments.

\subsection{Baselines}
We verify the efficacy of our approach (MoE) in \textit{non-adversarial} and \textit{adversarial} settings respectively. In both settings, we compare our approach against the following two baselines:
\begin{itemize}
    \item \textbf{best-SS}: the best single-source adaptation model among all the sources.
    \item \textbf{uni-MS}: the unified multi-source adaptation model, which is trained using the combination of all the source domain data with single-source transfer methods. uni-MS is a common and strong baseline for multi-source domain adaptation \cite{zhao2018multiple}.
\end{itemize}

For the rest of the paper, we name the adversarial counterpart of the models as \textbf{$*$-A}.


In the adversarial setting on \textsc{Chen12}, in addition to best-SS and uni-MS with adversarial loss, we further compare with the following two systems that also utilize unlabeled data from target domain.
\begin{itemize}
    \item \textbf{mSDA}: the marginalized stacked denoising autoencoder \cite{Chen2012msda}. mSDA outperforms prior deep learning and shallow learning approaches such as structural correspondence learning \cite{blitzer2007da} and denoising autoencoder \cite{glorot2011domain}.
    \item \textbf{MDAN}: the multi-source domain adversarial network \cite{zhao2018multiple}. MDAN gives the state-of-the-art performance for multi-source domain adaptation on \textsc{Chen12}. It generalizes the domain adversarial network to multiple source domain adaptation by selectively backpropagating the domain discrimination loss according to domain classification error.
\end{itemize}

\subsection{Implementation Details}
\label{subsec:implement}
For \textsc{Chen12}, since the dataset is in TF-IDF format and the word ordering information is not available, we use a multilayer perceptron with an input layer of 5,000 dimensions and one hidden layer of 500 dimensions as our encoder. For \textsc{Ziser17}, we instead use a convolutional neural network encoder with a combination of kernel widths 3 and 5 \cite{kim2014cnn}, each with one hidden layer of size 150, which are then concatenated to a 300 dimension representation.\footnote{Note that with a more extensive architecture search, we are likely to achieve better results. This, however, is not the main focus of this work.}

For the POS tagging encoder, we use a hierarchical bidirectional LSTM (BiLSTM) network, which contains a character-level BiLSTM for generating individual word representations, followed by a word-level BiLSTM that generates contextualized word representations.


For MMD, we follow \newcite{bousmalis2016domain} and use 19 RBF kernels with the standard deviation parameters ranging from $10^{-6}$ to $10^6$.\footnote{Detailed values are presented in the supplementary material in \newcite{bousmalis2016domain}.}


\begin{table*}[t]
    \centering
    \begin{tabular}{lcccccccc}
    \toprule
    \multirow{2}{*}{\textsc{Setting}} & \multicolumn{3}{c}{\textsc{Non-adversarial}} & \multicolumn{5}{c}{\textsc{adversarial}} \\
    \cmidrule(lr){2-4}\cmidrule(lr){5-9}
    & best-SS & uni-MS & MoE & mSDA$^\dagger$ & MDAN & best-SS-A & uni-MS-A & MoE-A \\
    \midrule
    D,E,K--B & 75.43 & 78.43 & \bf 79.42 & 76.98 & 78.63 & 80.07 & 80.25 & \bf 80.87 \\
    B,E,K--D & 81.23 & 82.49 & \bf 83.35 & 78.61 & 80.65 & 82.68 & 83.30 & \bf 83.99 \\
    B,D,K--E & 85.51 & 84.79$^*$ & \bf 86.62 & 81.98 & 85.34 & 86.32 & 85.96$^*$ & \bf 86.38 \\
    B,D,E--K & 86.83 & 87.00 & \bf 87.96 & 84.33 & 86.26 & 87.05 & 87.55 & \bf 88.06 \\
    \midrule
    \it Average & \it 82.25 & \it 83.18 & \it{\textbf{84.34}} & \it 80.48 & \it 82.72 & \it 84.03 & \it 84.27 & \it{\textbf{84.83}} \\
    \bottomrule
    \end{tabular}
    \caption{Multi-Source domain adaptation accuracy on Amazon dataset of \textsc{Chen12}. $^*$ indicates negative transfer, i.e., the unified multi-source model underperforms the best single-source model. mSDA$^\dagger$ is not an adversarial approach, but utilizes unlabeled data from target domain.}
    \label{tab:result-chen12}
\end{table*}

\begin{table*}[htbp]
    \centering
    \begin{tabular}{lcccccc}
    \toprule
    \multirow{2}{*}{\textsc{Setting}} & \multicolumn{3}{c}{\textsc{Non-adversarial}} & \multicolumn{3}{c}{\textsc{adversarial}} \\
    \cmidrule(lr){2-4}\cmidrule(lr){5-7}
    & best-SS & uni-MS & MoE & best-SS-A & uni-MS-A & MoE-A \\
    \midrule
    D,E,K--B     & 85.35 & 87.00 & \bf 87.55 & 86.85 & 87.55 & \bf 87.85 \\
    B,E,K--D     & 85.25 & 86.80 & \bf 87.85 & 86.00 & 87.40 & \bf 87.65 \\
    B,D,K--E     & 86.80 & 88.30 & \bf 89.20 & 88.90 & 89.35 & \bf 89.50 \\
    B,D,E--K     & 88.90 & 89.65 & \bf 90.45 & 89.95 & 90.35 & \bf 90.45 \\
    \midrule
    \it Average     & \it 86.58 & \it 87.94 & \it{\textbf{88.76}} & \it 87.93 & \it 88.66 & \it{\textbf{88.86}} \\
    \bottomrule
    \end{tabular}
    \caption{Multi-Source domain adaptation accuracy on Amazon dataset of \textsc{Ziser17}.}
    \label{tab:result-ziser17}
\end{table*}

\begin{figure*}[htbp]
    \centering
    \includegraphics[width=1.0\textwidth]{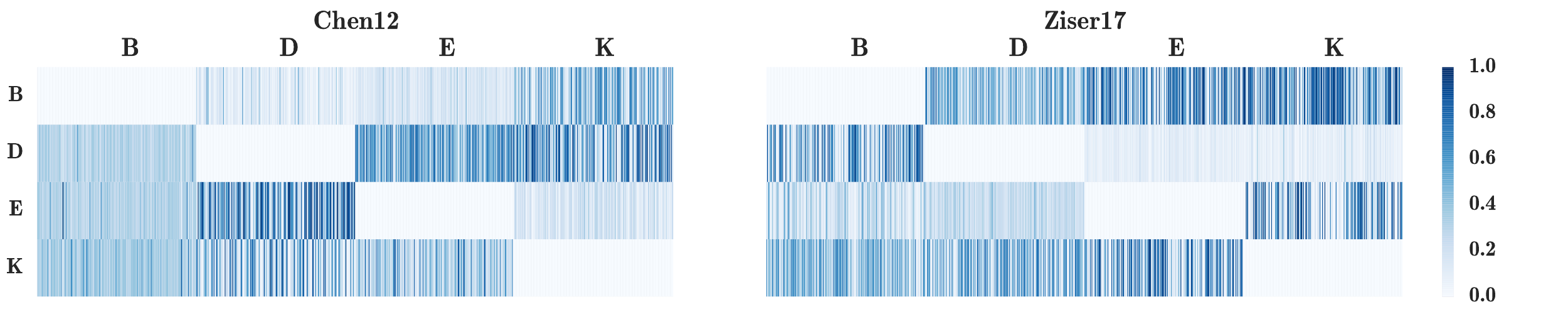}
    \vspace{-1.5em}
    \caption{$\alpha$ distributions across source domains for randomly selected 200 examples in each target domain of \textsc{Chen12} (left) and \textsc{Ziser17} (right). Columns represent target domains and rows represent sources.}
    \label{fig:alpha-amazon}
\end{figure*}

\begin{figure*}[htbp]
    \centering
    \includegraphics[width=1.0\textwidth]{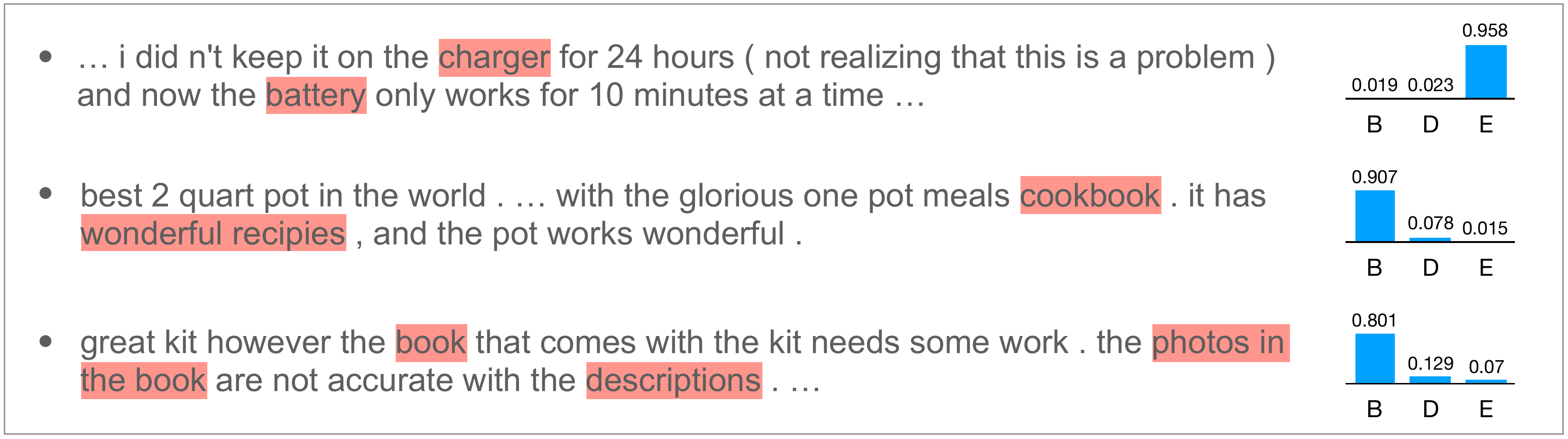}
    \vspace{-1.5em}
    \caption{Examples of \textit{Kitchen} (K) reviews in \textsc{Ziser17} and their $\alpha$ distribution over \textit{Books} (B), \textit{DVDs} (D) and \textit{Electronics} (E). The manually highlighted words indicate the specific Kitchen products described in the reviews.}
    \label{fig:alpha-example}
\end{figure*}

All the models were trained using Adam with weight decay. Learning rate is set to $10^{-4}$ for \textsc{Chen12} and $10^{-3}$ for \textsc{Ziser17} and POS tagging. We use mini-batches of 32 samples from each domain. We tune the coefficients $\lambda$, $\eta$ for each adaptation task. $\gamma$ is set to 1 for all experiments.

\section{Results}
\subsection{Sentiment Analysis on Amazon Reviews}

We report our results on the Amazon reviews datasets in Table \ref{tab:result-chen12} (\textsc{Chen12}) and Table \ref{tab:result-ziser17} (\textsc{Ziser17}).
Our approach (MoE) consistently achieves the best performance across different settings and tasks.

The results clearly demonstrate the value of using multiple sources.
In most cases, even a unified model performs better than the oracle best single source. 
By smartly combining all the sources, our model outperforms the unified model significantly.
One exception is the task of \textquotedblleft B,D,K-E\textquotedblright~in \textsc{Chen12}, where the unified multi-source model doesn't improve over the best single source model, constituting a \textit{negative transfer} scenario. However, even in this scenario, our approach still performs significantly better, demonstrating its robustness in handling \textit{negative transfer}.



\paragraph{Impact of adversarial adaptation}
We achieve consistent improvements over the baseline systems with the addition of the adversarial loss. In most cases, MoE also achieves additional improvement (e.g., 79.42\% \textit{vs.} 80.87\% in ``D,E,K-B"). We notice that in some cases, e.g., ``B,D,K-E" in \textsc{Chen12} and ``B,E,K-D" in \textsc{Ziser17}, the adversarial loss doesn't help MoE. This might be attributed to the fact that by aligning the target distribution with the source domains, the representation space becomes more compact, thus making it more difficult to capture source domain-specific characteristics and increasing the difficulty of metric learning in MoE.

\paragraph{Analysis on the metric ($\alpha$)}
Figure~\ref{fig:alpha-amazon} visualizes the distribution of $\alpha$ values, learned by our model in different tasks, across the source domains. The visualization is based on 200 examples for each domain randomly sampled from the corresponding test set. From the heatmap we can see that for a specific target domain, different examples may have different $\alpha$ distributions. Moreover, for most examples, the $\alpha$ distribution is skewed, indicating that our model draws on a few most informative source domains.

Figure~\ref{fig:alpha-example} exemplifies the above point. For instance, the first review about ``charger" and ``battery" is closer to the \textit{Electronics} source domain. This relation is successfully captured by the $\alpha$ distribution produced by our model.


\begin{table}[t]
    \centering
    \begin{tabular}{lcccc}
    \toprule
    \multirow{2}{*}{\textsc{Setting}} & \multicolumn{2}{c}{\small \textsc{Chen12} (w/o $\mathcal{R}_h$)} & \multicolumn{2}{c}{\small \textsc{Ziser17} (w/o $\mathcal{R}_h$)} \\
    \cmidrule(lr){2-3}\cmidrule(lr){4-5}
    & MoE & MoE-A & MoE & MoE-A \\
    \midrule
    D,E,K--B & \it -0.70 & \it -0.51 & \it -0.75 & \it -0.60 \\
    B,E,K--D & \it -0.67 & \it -0.41 & \it -0.05 & \it -1.20 \\
    B,D,K--E & \it -1.93 & \it -0.44 & \it -0.70 & \it -0.60 \\
    B,D,E--K & \it -0.49 & \it -0.09 & \it -0.50 & \it +0.30 \\
    \bottomrule
    \end{tabular}
    \caption{Ablation test of entropy regularizer on \textsc{Chen12} and \textsc{Ziser17} (decrease in accuracy).}
    \vspace{-0.5em}
    \label{tab:ablation-entropy}
\end{table}

We further investigate the impact of entropy regularization over $\alpha$.
Table \ref{tab:ablation-entropy} summarizes the ablation test results of entropy regularization ($\eta = 0$) on \textsc{Chen12} and \textsc{Ziser17}. It shows that entropy regularization benefits our model under both non-adversarial and adversarial settings.

\begin{table*}[htbp]
    \centering
    \begin{tabular}{lcccccccc}
    \toprule
    \multirow{2}{*}{\textsc{Target}} & \multicolumn{4}{c}{\textsc{Non-adversarial}} & \multicolumn{4}{c}{\textsc{adversarial}} \\
    \cmidrule(lr){2-5}\cmidrule(lr){6-9}
    & best-SS & uni-MS & uni-MS$^\dagger$ & MoE & best-SS-A & uni-MS-A & uni-MS-A$^\dagger$ & MoE-A \\
    \midrule
    Answers   & 88.16 & 88.89 & 89.88 & \bf 90.26 & 88.47 & 89.04 & \bf 89.99 & \bf 89.80 \\
    Reviews   & 87.15 & 87.45 & 88.91 & \bf 89.37 & 87.26 & 87.90 & 88.94 & \bf 89.40 \\
    Newsgroup & 89.14 & 89.95 & 90.70 & \bf 91.03 & 89.54 & 90.20 & 90.70 & \bf 91.13 \\
    \midrule
    \it Average   & \it 88.15 & \it 88.76 & \it 89.83 & \it{\textbf{90.22}} & \it 88.42 & \it 89.05 & \it 89.88 & \it{\textbf{90.11}} \\
    \bottomrule
    \end{tabular}
    \caption{POS tagging results on SANCL data. Source domains include \textit{Web}, \textit{Emails}, \textit{Twitter}. $^\dagger$ indicates the unified multi-source model trained without \textit{Twitter}, thus can be considered as the oracle performance (upper-bound) of uni-MS.}
    \label{tab:pos-tagging}
\end{table*}


\subsection{Part-of-Speech Tagging}
\label{subsec:pos-results}

Table \ref{tab:pos-tagging} summarizes our results on POS tagging. Again, our approach consistently achieves the best performance across different settings and tasks.
Adding \textit{Twitter} as a source leads to a drop in performance for the unified model, as a result of negative transfer. Our method, however, robustly handles negative transfer and manages to even benefit from this additional source. 



\paragraph{Impact of negative transfer}
Table \ref{tab:pos-tagging-alphas} presents the $\alpha$ distribution learned by the metric, on average for all tokens of the target domain. As we can see, our model (MoE-A) effectively learns to decrease the weights on \textit{Twitter}, demonstrating again its ability to alleviate negative transfer.

\begin{table}[htbp]
    \centering
    \begin{tabular}{lccc}
    \toprule
    \multirow{2}{*}{\textsc{Target}} & \multicolumn{3}{c}{\textsc{Source}} \\
    \cmidrule(lr){2-4}
    & Twitter & Emails & Web \\
    \midrule
    Answers   & 0.0527 & 0.5941 & 0.3531 \\
    Reviews   & 0.0640 & 0.5250 & 0.4100 \\
    Newsgroup & 0.0538 & 0.4960 & 0.4490 \\
    \bottomrule
    \end{tabular}
    \caption{Distribution of the metric values $\alpha$ on average for all tokens in the SANCL test set.} 
    \label{tab:pos-tagging-alphas}
\end{table}

We further study the impact of this outlier source by varying the amount of \textit{Twitter} data used during training.
We gradually increase the number of \textit{Twitter} instances by 750. As shown in Table~\ref{tab:pos-varying-twitter}, the increase of the \textit{Twitter} data does not benefit the unified multi-source model (uni-MS-A), and even amplifies negative transfer for the \textit{Answers} and \textit{Reviews} domains. However, the performance of our MoE (MoE-A) model stays stable, consistently increasing with more \textit{Twitter}, showing robustness in handling negative transfer.

\begin{table}[h]
    \centering
    \begin{tabular}{llccc}
    \toprule
    \multirow{2}{*}{\textsc{Target}} & \multirow{2}{*}{\specialcell{\textsc{Model}\\~~(\textbf{$*$-A})}} & \multicolumn{3}{c}{\# Twitter instances} \\
    \cmidrule(lr){3-5}
    &  & 750 & 1,500 & 2,250 \\
    \midrule
    \multirow{2}{*}{Answers}    & uni-MS & 89.04 & 89.04 & 86.93 \\
                                & MoE    & 89.80 & 91.22 & 90.90 \\
    \cmidrule(lr){2-5}
    \multirow{2}{*}{Reviews}    & uni-MS & 87.90 & 87.45 & 87.68 \\
                                & MoE    & 89.40 & 90.23 & 91.14 \\
    \cmidrule(lr){2-5}
    \multirow{2}{*}{Newsgroup}  & uni-MS & 90.20 & 90.10 & 90.21 \\
                                & MoE    & 91.13 & 91.32 & 91.82 \\
    \bottomrule
    \end{tabular}
    \caption{POS tagging accuracy with varying amounts of \textit{Twitter} data in training.}
    \label{tab:pos-varying-twitter}
\end{table}

\section{Conclusion}
In this paper, we propose a novel mixture-of-experts (MoE) approach for unsupervised domain adaptation from multiple diverse source domains. We model the domain relations through a \textit{point-to-set} distance metric, and introduce a \textit{meta-training} mechanism to learn this metric. Experimental results on sentiment classification and part-of-speech tagging demonstrate that our approach consistently outperforms various baselines and can robustly handle negative transfer. The effectiveness of our approach suggests its potential application to a broader range of domain adaptation tasks in NLP and other areas.




\section*{Acknowledgments}
We thank MIT NLP group and the anonymous reviewers for their helpful comments. We also thank Shiyu Chang and Mo Yu for insightful discussions on metric learning. This work is supported by the MIT-IBM Watson AI Lab. Any opinions, findings, conclusions, or recommendations expressed in this paper are those of the authors, and do not necessarily reflect the views of the funding organizations.

\bibliography{acl2018}
\bibliographystyle{acl_natbib}

\appendix


\end{document}